\documentclass[journal]{IEEEtran}

\ifCLASSINFOpdf
\else
\fi

\usepackage{graphicx}
\usepackage{amsmath}
\usepackage{cite}
\usepackage{enumerate}
\usepackage{cases}
\usepackage{multirow}
\usepackage{verbatim}
\usepackage{amssymb}
\usepackage{color}

\usepackage{bm}
\usepackage{booktabs}
\usepackage{hyperref}
\usepackage{CJK}
\usepackage{algorithm}
\usepackage{algorithmicx}
\usepackage{algpseudocode}

\floatname{algorithm}{Algorithm}

\begin{document}

\title{An Iterative Co-Saliency Framework for RGBD Images}


\author{Runmin Cong, Jianjun Lei,~\IEEEmembership{Senior Member,~IEEE,} Huazhu Fu, Weisi Lin,~\IEEEmembership{Fellow,~IEEE,}\par
 Qingming Huang,~\IEEEmembership{Senior Member,~IEEE,} Xiaochun Cao,~\IEEEmembership{Senior Member,~IEEE,} and Chunping Hou
\thanks{Manuscript received Janurary 31, 2017; revised July 09, 2017; accepted October 23, 2017. This work was supported in part by the National Natural Science Foundation of China under Grant 61722112, Grant 61520106002, Grant 61731003, Grant 61332016, Grant 61620106009, Grant U1636214, Grant 61602344, and in part by the National Key R\&D Program of China under Grant 2017YFB1002900. (\emph{Corresponding author: Jianjun Lei.})}
\thanks{R. Cong is with the School of Electrical and Information Engineering, Tianjin University, Tianjin 300072, China, and also with the School of Computer Engineering, Nanyang Technological University, Singapore 639798 (e-mail: rmcong@tju.edu.cn).}
\thanks{J. Lei, and C. Hou are with the School of Electrical and Information Engineering, Tianjin University, Tianjin 300072, China (e-mail: jjlei@tju.edu.cn; hcp@tju.edu.cn).}
\thanks{H. Fu is with the Ocular Imaging Department, Institute for Infocomm Research, Agency for Science, Technology and Research, Singapore 138632 (e-mail: huazhufu@gmail.com).}
\thanks{W. Lin is with the School of Computer Engineering, Nanyang Technological University, Singapore 639798 (e-mail: wslin@ntu.edu.sg).}
\thanks{Q. Huang is with the School of Computer and Control Engineering, University of Chinese Academy of Sciences, Beijing 100190, China (e-mail: qmhuang@ucas.ac.cn).}
\thanks{X. Cao is with State Key Laboratory of Information Security, Institute of Information Engineering, Chinese Academy of Sciences, Beijing 100093, China, and also with University of Chinese Academy of Sciences (e-mail: caoxiaochun@iie.ac.cn).}
}

\markboth{IEEE TRANSACTIONS ON CYBERNETICS, ~Vol.~xx, No.~xx, xxxx~2017}%
{Shell \MakeLowercase{\textit{et al.}}: Bare Demo of IEEEtran.cls for IEEE Journals}

\maketitle

\begin{abstract}
As a newly emerging and significant topic in computer vision community, co-saliency detection aims at discovering the common salient objects in multiple related images. The existing methods often generate the co-saliency map through a direct forward pipeline which is based on the designed cues or initialization, but lack the refinement-cycle scheme. Moreover, they mainly focus on RGB image and ignore the depth information for RGBD images. In this paper, we propose an iterative RGBD co-saliency framework, which utilizes the existing single saliency maps as the initialization, and generates the final RGBD co-saliency map by using a refinement-cycle model. Three schemes are employed in the proposed RGBD co-saliency framework, which include the addition scheme, deletion scheme, and iteration scheme. The addition scheme is used to highlight the salient regions based on intra-image depth propagation and saliency propagation, while the deletion scheme filters the saliency regions and removes the non-common salient regions based on inter-image constraint. The iteration scheme is proposed to obtain more homogeneous and consistent co-saliency map. Furthermore, a novel descriptor, named depth shape prior, is proposed in the addition scheme to introduce the depth information to enhance identification of co-salient objects. The proposed method can effectively exploit any existing 2D saliency model to work well in RGBD co-saliency scenarios. The experiments on two RGBD co-saliency datasets demonstrate the effectiveness of our proposed framework.
\end{abstract}


\IEEEpeerreviewmaketitle

\section{Introduction}

\IEEEPARstart{H}{UMAN} visual system serves as a filter for selecting the salient and interesting regions for further processing. In computer vision community, saliency detection methods are proposed to simulate this characteristic of early primate visual system, which aim at capturing the most salient and informative regions in an image, and have been applied in a wide range of vision applications, such as image retrieval \cite{R1}, sensation enhancement \cite{R2,R2-2}, foreground annotation \cite{R3}, image segmentation \cite{R4,R4-1}, image quality assessment \cite{R5}, image retargeting \cite{R6,R7}, and coding \cite{R7-2}. Numerous saliency detection models \cite{R8,R9,R10,R10-2,R11,R12,R13} focus on detecting the salient objects from a single image, which achieve encouraging performance on the public benchmarks.\par
\begin{figure}[!t]
\centering
\includegraphics[width=1\linewidth]{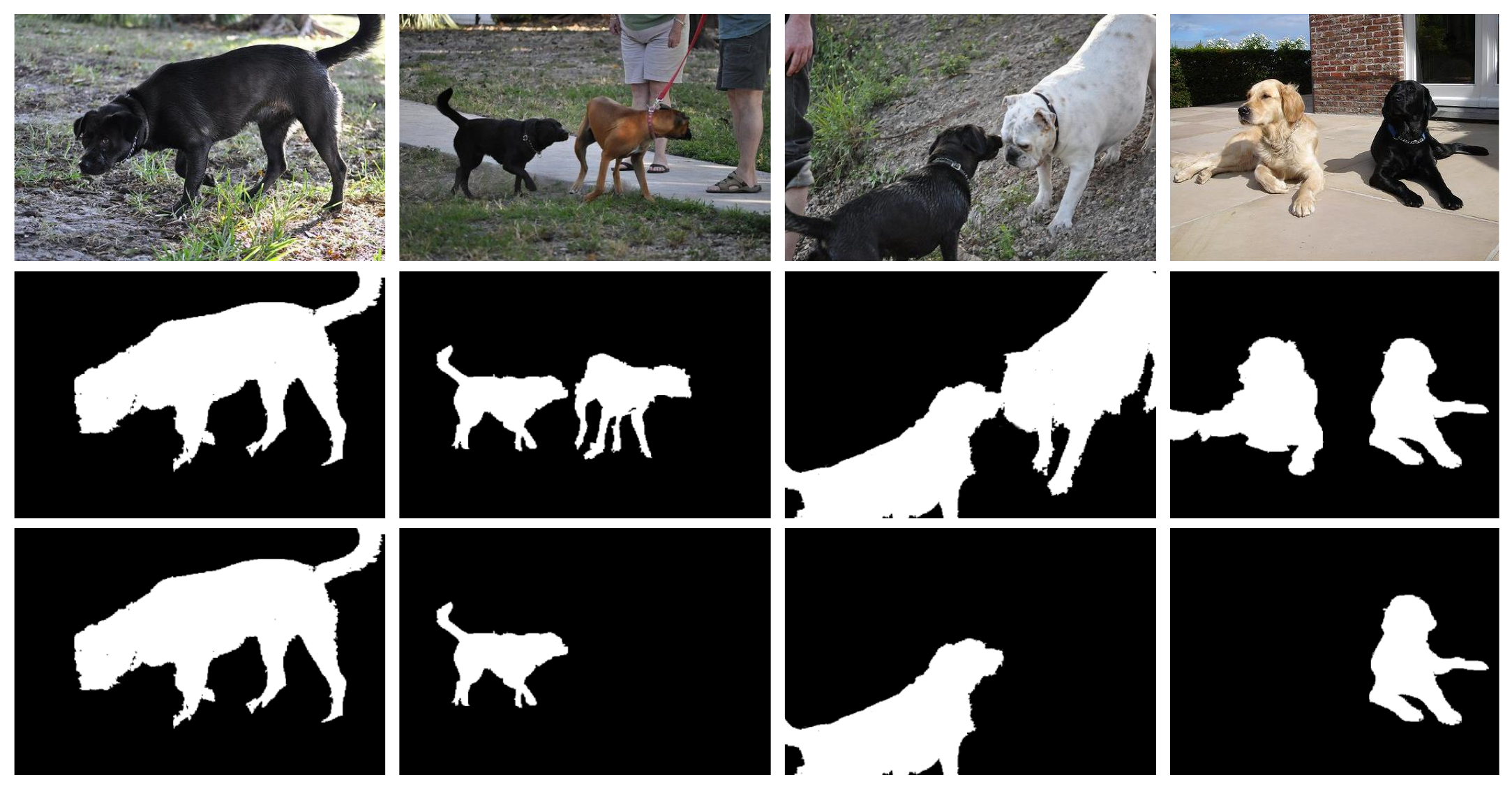}
\caption{An example of the co-saliency detection. The first row is the input RGB images in an image group, the second row shows some saliency detection results, and the third row represents some co-saliency detection results. From the perspective of the image group, only the black dog is the common salient object in the co-saliency detection model. }
\label{fig1}
\end{figure}

As an emerging and challenging issue, co-saliency detection has been attracting more attentions in recent years. Different from the traditional single saliency detection model, co-saliency detection methods focus on discovering the common salient objects in multiple images. Based on the definition of co-saliency \cite{R14,R15,R16,R17,R18}, two main properties of the co-salient object should be owned simultaneously, i.e., 1) the target objects should be salient in individual image, and 2) all co-salient objects should be common among multiple images. In fact, the categories, intrinsic characteristics and locations of these objects are entirely unknown, and the background of the scenes is different. Fig. \ref{fig1} provides an example of co-saliency detection. The two dogs in the last three images are salient in the single image saliency model. However, from the perspective of the image group, only the black dog is the common salient object in this image group. Therefore, co-saliency detection is a more challenging issue compared to saliency detection, while the extracted co-salient regions are more useful in many computer vision tasks, such as object co-localization \cite{R19,R20}, image matching \cite{R21}, foreground co-segmentation \cite{R22,R23}, and co-detection \cite{R24}. Most existing co-saliency detection models \cite{R25,R26,R27,R28,R29,R30,R31,R32,R33} focus on designing the complete and independent algorithms to discover the common salient regions and obtaining the satisfactory performances.\par
In fact, the single saliency map produced by the existing saliency model can be considered as an initialization in co-saliency detection. Moreover, the existing co-saliency detection methods mainly rely on the designed cues or initialization, and lack the refinement-cycle. In this paper, we propose an effective co-saliency framework based on the refinement-cycle model, which integrates the addition scheme, deletion scheme, and iteration scheme. The addition scheme is used to enrich the saliency regions through the depth propagation and saliency propagation. The inter saliency model is formalized as common probability calculation to capture the inter-image correspondence in the deletion scheme. Moreover, the iterative optimization scheme is designed to achieve more superior co-saliency result in our framework.\par

In addition, the depth information from the RGBD image has been demonstrated the usefulness for many computer vision tasks \cite{R34-2,R34-3}, and has been introduced into many saliency models to enhance the detection performance \cite{R34,R35,R36,R37,R38,R39}. Niu \emph{et al.} \cite{R34} used the global disparity contrast and domain knowledge to capture the depth information. Peng \emph{et al.} \cite{R35} calculated the depth saliency from multi-contextual contrast including the local context, global context and background context. Feng \emph{et al.} \cite{R37} proposed a local background enclosure (LBE) feature to evaluate the depth saliency. However, in the above methods, the information of the relevant and similar objects in a sequence of images is ignored and not exploited. In our paper, a novel depth descriptor, named depth shape prior, is proposed to capture the shape attributes from the depth map to improve the co-saliency detection performance.\par
In summary, most of existing co-saliency methods aim to design a single forward pipeline which generates the co-saliency map based on the designed cues directly, but lack the refinement-cycle scheme and ignore the depth information for RGBD images. Thus, in our work, we propose an iterative RGBD co-saliency framework, which utilizes the additional depth information and employs the existing RGB saliency map as the initialization in a refinement-cycle model to produce the final RGBD co-saliency map. The major contributions of the proposed method are:\par
\begin{enumerate}[1)]
\item An iterative co-saliency framework for RGBD images is proposed, which integrates addition scheme, deletion scheme, and iteration scheme. The intra-image propagations and the inter-image constraints are incorporated into a cyclic model to achieve the co-saliency detection.
\item A novel depth descriptor, named depth shape prior, is proposed to capture the shape attributes from the depth map and enhance the identification of co-salient objects from RGBD images.
\item A superpixel-level common probability function among multiple images is calculated to exploit the inter-image corresponding relationship in the deletion scheme.
\item An iterative updating strategy is designed to obtain more homogeneous and consistent co-saliency result in the iteration scheme.
\end{enumerate}\par
The rest of this paper is organized as follow. Section II reviews the related works of saliency and co-saliency detection briefly. Section III details the proposed RGBD co-saliency detection framework. The experimental results and analysis with quantitative evaluation are presented in Section IV. Finally, the conclusion is provided in Section V.\par
\section{Related Work}

In the last decade, a number of methods \cite{R8,R9,R10,R10-2,R11,R12,R13,R40,R41,R42,R43,R44} have been presented to identify the salient regions from a RGB image, which can be applied in many computer vision tasks. Cheng \emph{et al.} \cite{R8} proposed a regional contrast based saliency detection algorithm, which simultaneously evaluates global contrast differences and spatial weighted coherence scores. Li \emph{et al.} \cite{R10} proposed a novel graph-based saliency detection algorithm, which formulates the pixel-wised saliency maps using the regularized random walks ranking. Shi \emph{et al.} \cite{R11} designed a hierarchical saliency model to compute the saliency cues using weighted color contrast on three image layers. More recently, the theory of deep learning has been applied in saliency detection, and has achieved remarkable performance. In \cite{R41}, Chen \emph{et al.} built a saliency model with two stacked CNNs, and the coarse-level and fine-level representation learning are utilized to learn the saliency representation in a progressive manner. Lee \emph{et al.} \cite{R43} integrated the hand-crafted features and high-level features into the saliency model to enhance performance of saliency detection. To address the blurry boundary of the salient object, Li \emph{et al.} \cite{R44} proposed an end-to-end deep contrast network, which includes pixel-level fully convolutional stream and segment-wise spatial pooling stream.\par
In addition, human can perceive the surroundings by an additional depth cue that is provided by stereopsis, which plays an important role in the human visual system. The depth information has demonstrated its usefulness for many computer vision tasks \cite{R23,R35,R45}. Some saliency detection methods integrated with depth information, named RGBD saliency, are proposed in recent years. In \cite{R35}, considering low-level feature contrast, mid-level region grouping and high-level object-aware priors, Peng \emph{et al.} proposed a multi-stage RGBD model to discover the salient regions. Ju \emph{et al.} \cite{R36} proposed a depth-aware anisotropic center-surround difference (ACSD) measurement to evaluate the depth saliency. Feng \emph{et al.} \cite{R37} used the local background enclosure (LBE) feature to compute the RGBD saliency. Considering the quality of depth map, Cong \emph{et al.} \cite{R38} proposed a RGBD saliency model by combining depth confidence analysis and multiple cues fusion. In \cite{R39}, Sheng \emph{et al.} introduced the depth contrast increased to make the saliency analysis easier and more accurate.\par

\begin{figure*}[!t]
\centering
\includegraphics[width=1\linewidth]{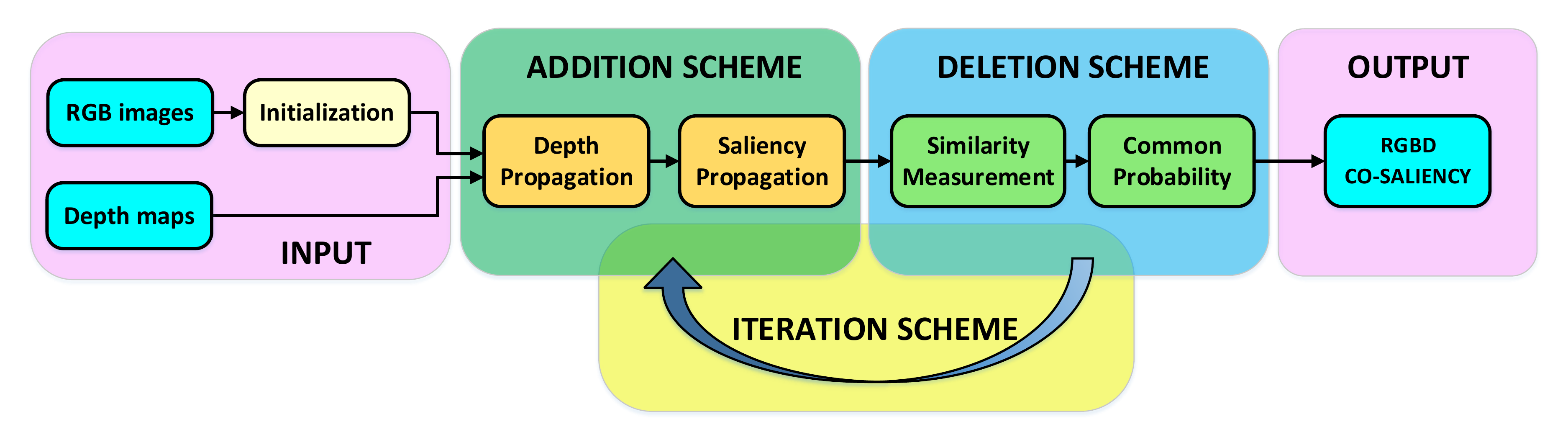}
\caption{The flowchart of the proposed RGBD co-saliency framework. }
\label{fig2}
\end{figure*}

Different from modeling the human visual mechanism from a single image, co-saliency detection makes effort to discover the common salient objects when people are reviewing a group of related images. There are many methods have been proposed to achieve the co-saliency detection. In \cite{R16}, the single-image and multi-image saliency maps are linearly combined to detect the co-salient regions. However, this method is only available for image pairs. In \cite{R18}, a cluster-level method that integrates five saliency cues is proposed to compute the inter saliency. Liu \emph{et al.} \cite{R25} proposed a co-saliency model based on hierarchical segmentation, which includes the fine-level and coarse-level segmentation. In \cite{R26}, co-saliency detection is formulated as a two-stage saliency propagation problem that uses the single-image saliency map to propagate pairwise saliency values. Cao \emph{et al.} \cite{R29} proposed a general fusion framework for saliency and co-saliency detection, which introduces the self-adaptively weighted scheme via rank constraint. Li \emph{et al.} \cite{R30} proposed a two-stage guided scheme to obtain the guided saliency map for each single image through ranking framework, and the final co-saliency map is produced by fusing these guided saliency maps. To detect the salient objects from complex natural scenes with small-scale high-contrast backgrounds, Huang \emph{et al.} \cite{R31} presented a saliency detection method via multiscale low-rank analysis, and introduced a GMM-based co-saliency detection method. Recently, the learning-based co-saliency models are drawing more attention from researchers due to its superior performance. Zhang \emph{et al.} \cite{R32} exploited convolutional neural network (CNN) with additional transfer layers to generate the higher-level features, and these features are used to discover the co-salient objects via Bayesian framework. In \cite{R33}, the co-saliency detection is formulated under a multiple-instance learning (MIL) framework, and self-paced learning (SPL) regime is introduced into the MIL framework for selecting training samples in a theoretically sound manner.\par

However, all the above-mentioned co-saliency models focus on RGB images, and ignore the effectiveness of depth information to enhance the identification of co-salient objects from the cluttered and changing natural scene. In \cite{R50}, a RGBD co-saliency model using bagging-based clustering is proposed. Our work is different with method \cite{R50} as: 1) The motivation is different. The method in \cite{R50} utilized the bagging-based clustering in a single forward pipeline, which proceeds from the RGBD saliency map. By contrast, our work aims at designing an iterative co-saliency framework, which includes addition, deletion, and iteration schemes to achieve the co-saliency detection. Moreover, we employ the 2D saliency map as the initial map, and incorporate the additional depth information in the model. Thus, our method can convert any 2D saliency map into the RGBD co-saliency map. 2) The use of depth information is different. The method in \cite{R50} focuses on the basic depth distribution features, \emph{e.g.}, depth value, depth range, and HOG histogram on the depth map. By contrast, a novel depth descriptor named depth shape prior is proposed in our method, which captures the shape attributes from the depth map and improves the performance of the co-saliency detection by using the depth consistency and shape attributes.
 \par

\section{Proposed Method}

\begin{figure}[!t]
\centering
\includegraphics[width=1\linewidth]{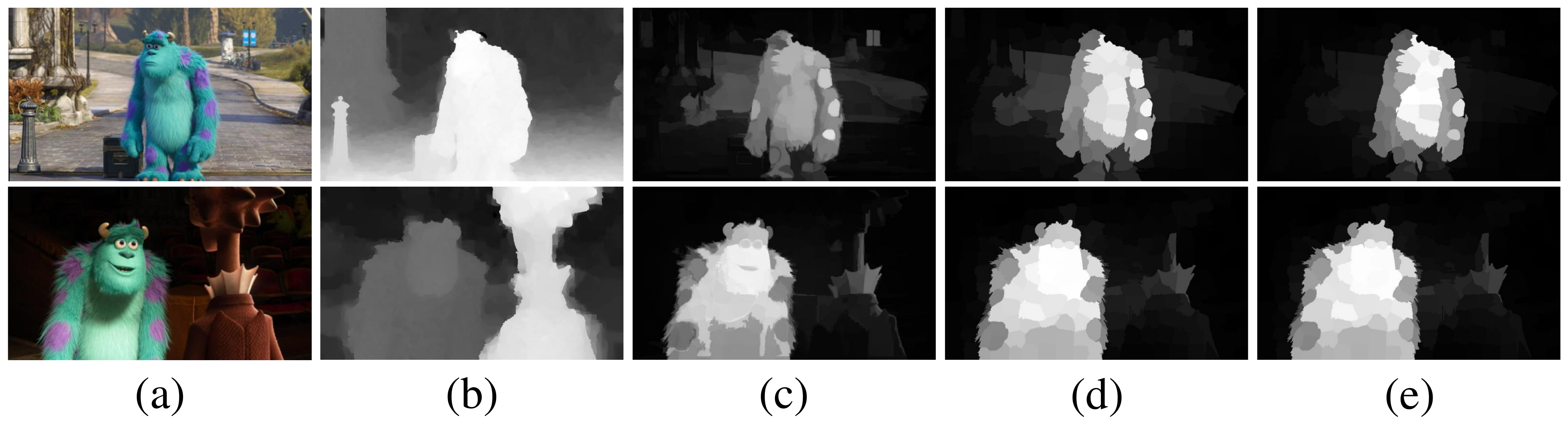}
\caption{Some examples of the proposed method. (a) RGB image. (b) Depth map. (c) The initialized saliency map. (d) The co-saliency map without iteration. (e) The final co-saliency map with iteration. }
\label{fig3}
\end{figure}\par

The proposed RGBD co-saliency framework is introduced in this section. Fig. \ref{fig2} shows the framework of the proposed method. Our method is initialized by the existing 2D saliency maps, and then three schemes are employed to generate the final RGBD co-saliency map. The addition scheme is used to grow the initialized saliency map from the perspective of intra-image, the deletion scheme is designed to suppress the non-common regions from the perspective of inter-image, and the iteration scheme is exploited to obtain more homogeneous and consistent co-saliency map. Some visual examples of the proposed method are shown in Fig. \ref{fig3}. \par

\textbf{Notations:} Given $N$ input RGB images $\{I^{i}\}_{i=1}^{N}$, and the corresponding depth maps are denoted as $\{D^{i}\}_{i=1}^{N}$. The $M_i$ single saliency maps for image $I^{i}$ produced by existing single image saliency models are represented as $S^{i}=\{S^{i}_j\}_{j=1}^{M_i}$. In our method, the superpixel-level region is regarded as the basic unit for processing. Thus, each RGB image $I^{i}$ is abstracted into superpixels $R^{i}=\{r_{m}^{i}\}_{m=1}^{N_{i}}$ using SLIC algorithm \cite{R46} firstly, where $N_{i}$ is the number of superpixels for image $I^{i}$.\par

\subsection{Initialization}
The proposed co-saliency framework aims at discovering the co-salient objects from multiple images in a group with the assistance of existing 2D saliency maps. Therefore, some existing saliency maps produced by 2D saliency models are used to initialize the framework. It is well known that different saliency methods own different superiority in detecting salient regions. In a way, these saliency maps are complementary in some regions, thus, the fused result can inherit the merits of the multiple saliency maps, and produce more robust and superior detection baseline. In our method, the simple average function is used to achieve a more generalized initialization result. The initialized saliency map for image $I^i$  is denoted as:
\begin{equation}
S_{f}^{i}(r^i_m)=\frac{1}{M_i}\sum_{j=1}^{M_i} S_{j}^{i}(r^i_m)
\end{equation}
where $S_{j}^{i}(r^i_m)$ denotes the saliency value of superpixel $r^i_m$ produced by $j^{th}$ saliency method for image $I^i$, and $M_i$ is the number of saliency maps for image $I^i$. In our experiments, five saliency methods including RC \cite{R8}, DCLC \cite{R9}, RRWR \cite{R10}, HS \cite{R11}, and BSCA \cite{R12}, are used to produce the 2D initialized saliency map. Some examples of the initialized saliency map are shown in Fig. \ref{fig3}(c). From the figures, we can see that the initialized result produces an impressive baseline for later co-saliency detection.\par

\subsection{Addition Scheme}
In this section, the addition scheme is designed to extend the saliency region based on the intra-image constraint by using two propagation algorithms. Firstly, a novel depth descriptor, named depth shape prior, is proposed to capture the depth cue and produce a RGBD saliency result in depth propagation. Then, saliency propagation is utilized to optimize and improve the saliency result furtherly. \par

\subsubsection{\textbf{Depth propagation}}
After initialization, the merits of the different saliency maps are inherited into the initialized saliency map. The depth information is introduced into the framework to enhance the identification of salient objects due to its usefulness in saliency detection. In general, the depth map owns the following properties: i) the salient object appears higher depth value compared to the backgrounds, ii) the high quality depth map can provide sharp and explicit boundary of the object, and iii) the interior depth value of the object should be smoothness and consistency. Inspired by these observations, a depth descriptor, namely depth shape prior (DSP), is proposed to capture the shape attributes from the depth map and improve the performance of the co-saliency detection by using the depth consistency and shape attributes. The proposed DSP descriptor is based on depth propagation and region grow. Several identified superpixels are selected as the seeds firstly, and then the DSP map can be calculated via depth constraints.\par
For each image $I^i$, the top $K$ superpixels with higher initialized saliency value are selected as the root seeds, which is represented as $\{r^i_{rk}\}^K_{k=1}$, and the corresponding depth shape prior map $DSP^i_k$ is initialized as zero.\par
For each root seed, we determine a set of child nodes $\{r^i_{cp}\}$ to depict the depth shape based on the depth smoothness and consistency constraints. In the $l$-loop diffusion, the superpixels direct neighboring the $(l-1)$-loop child nodes are selected as the $l$-loop child nodes only if they satisfy the following two constraints: \par
\begin{enumerate}[(a)]
\item \emph{Depth smoothness}: the depth difference between the neighbor superpixel and  $(l-1)$-loop child seeds is less than a certain threshold  $T_1$, as $|d^i_{nq}-d^i_{c,l-1}|\leq T_1$, where $d^i_{nq}$ is the depth value of the neighbor superpixel $r^i_{nq}$, and $d^i_{c,l-1}$ is the average depth value of $(l-1)$-loop child seeds.
\item \emph{Depth consistency}: the depth difference between the neighbor superpixel and root seed should be smaller than a specific threshold  $T_2$, as $|d^i_{nq}-d^i_{rk}|\leq T_2$, where $d^i_{rk}$ is the depth value of the root seed $r^i_{rk}$.
\end{enumerate}\par

Be noted that the child node in the first loop diffusion is initialized by the root seed in our method, and the two thresholds are set to 0.1 and 0.2 respectively. The depth shape prior value of the child node $r^i_{cp}$ in the $l$-loop is defined as:
\begin{equation}
DSP^i_k(r^i_{cp})=1-\min(|d^i_{cp,l}-d^i_{c,l-1}|,|d^i_{cp,l}-d^i_{rk}|)
\end{equation}
where $d^i_{cp,l}$ is the depth value of the child node $r^i_{cp}$ in the $l$-loop,  $d^i_{c,l-1}$ is the average depth value of $(l-1)$-loop child node set, and $|\cdot |$ is the absolute value function. Then, the next loop diffusion will be continued until there is no neighboring superpixel satisfies the depth constraints.\par
In our method, the top $K$ root seeds are selected for each image $I^i$ to improve the robustness, and $K$ DSP value maps are obtained for each image. Therefore, the final DSP map is defined as:
\begin{equation}
DSP^{i}(r^i_m)=\frac{1}{K}\sum_{k=1}^{K} DSP_{k}^{i}(r^i_m)
\end{equation}
where $K$ is the number of the root seeds, which is fixed to 10 in all the experiments.\par
\begin{figure}[!t]
\centering
\includegraphics[width=1\linewidth]{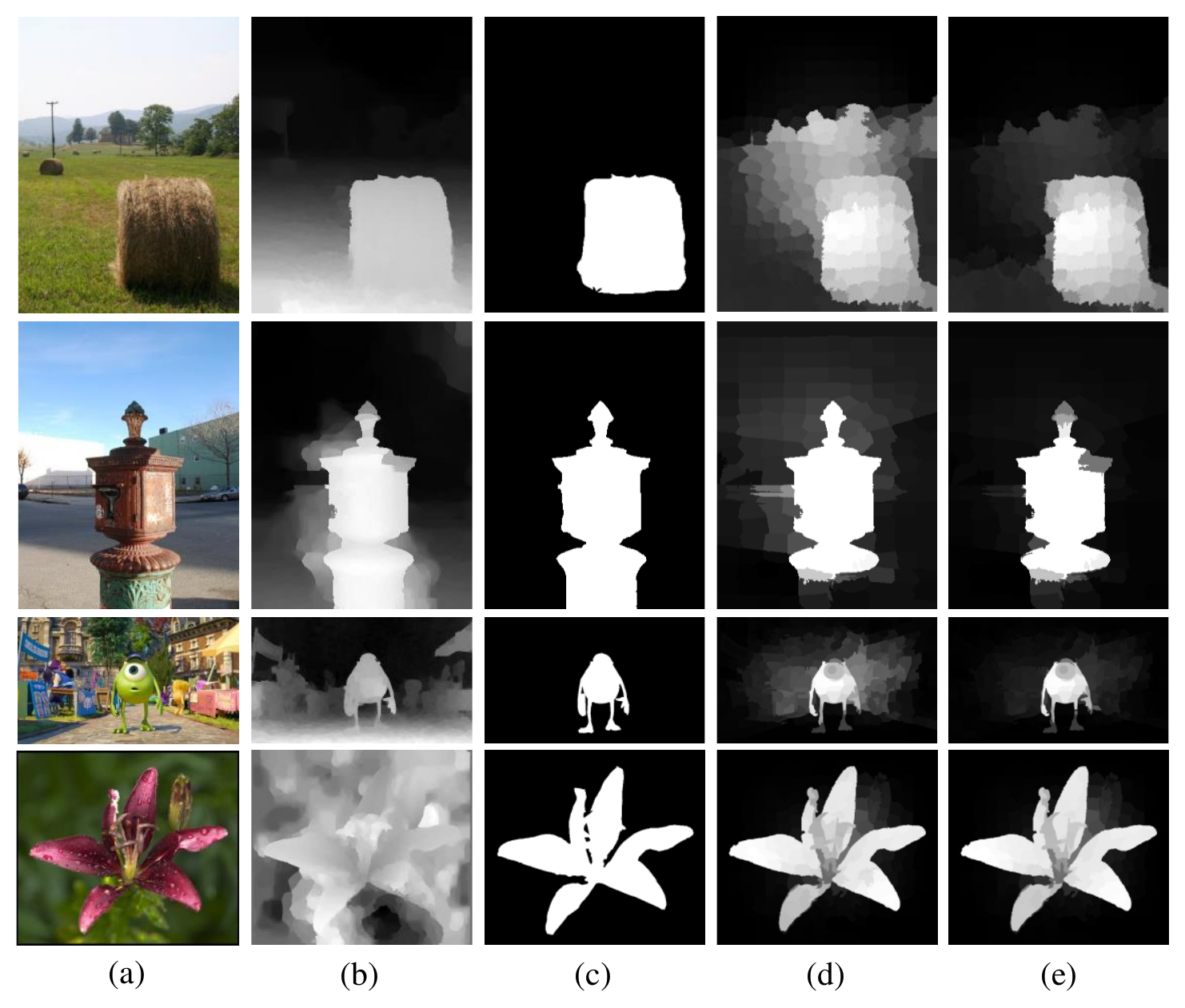}
\caption{The examples of the depth propagation. (a) RGB image. (b) Depth map. (c) Ground truth. (d) The RGB saliency result. (e) The RGBD saliency result with DSP descriptor.}
\label{fig4}
\end{figure}\par

To achieve more superior and stable saliency result, the initialized RGB saliency and the DSP map are combined in our method. Because the bad depth map may degenerate the accuracy of DSP map generation, we introduce the depth confidence measure $\lambda_d$ to evaluate the quality of the depth information, which is defined as \cite{R38}:
\begin{equation}
\lambda_{d}=\exp((1-m_{d})\cdot\emph{CV}\cdot\emph{H})-1
\end{equation}
where $m_{d}$ is the mean value of the whole depth image, $\emph{CV}$ denotes coefficient of variation, and $\emph{H}$ is the depth frequency entropy, which denotes the randomness of depth distribution. A larger $\lambda_d$ value represents greater reliability of the input depth map. More details of depth confidence measure can be found in \cite{R38}. Thus, the RGBD saliency that integrates initialized saliency map and DSP map weighted by depth confidence measure according to the depth quality is defined as:
\begin{equation}
S_{dp}^{i}(r^i_m)=(1-\lambda^i_d) \cdot S_{f}^{i}(r^i_m)+ \lambda^i_d\cdot S_{f}^{i}(r^i_m) \cdot DSP^{i}(r^i_m)
\end{equation}
where $\lambda^i_d$ is the depth confidence measure for image $I^i$, $S_{f}^{i}(r^i_m)$ represents the initialized saliency value of superpixel $r^i_m$, and $DSP^{i}(r^i_m)$ is the depth shape prior value of superpixel $r^i_m$. The obtained saliency map is normalized into [0,1]. With this depth confidence measure, the poor-quality depth map will be limited in the combination with RGB feature to avoid the degradation of the RGBD co-saliency result. Fig. \ref{fig4} shows some examples of the depth propagation. Comparing with the RGB saliency maps, some background regions around the salient object are suppressed effectively through depth propagation, such as the lawns in the first image, the roads in the second image, and the buildings in the third image. Moreover, the RGBD saliency model is more robust. Even if the quality of depth map is bad, such as the last raw in Fig. \ref{fig4}, our model still achieves better result by highlighting the RGB saliency component while DSP descriptor can not exploit accurate shape attributes from the poor-quality depth map.\par

\subsubsection{\textbf{Saliency propagation}}
With the obtained RGBD saliency map, the saliency propagation is conducted to further optimize the result. In our method, the superpixels are classified into three groups based on the saliency value firstly, which is denoted as the saliency seed superpixels, background seed superpixels, and the unknown superpixels. Then, saliency propagation is used to propagate the saliency of unknown superpixels on the graph from the saliency and background seeds.\par
For image $I^i$, a graph $G^i=(\vartheta^i,\varepsilon^i)$ among superpixels is constructed firstly, where $\vartheta^i$ denotes the node set which corresponds to the superpixels, and $\varepsilon^i$ is the link set among adjacent nodes. The affinity matrix $\textbf{\emph{W}}^{i}=[w_{uv}^{i}]_{N_{i}\times N_{i}}$ is defined as the similarity between two adjacent superpixels.
\begin{equation}
w_{uv}^{i}=
\begin{cases}
\exp(-\frac{\|\textbf{\emph{c}}_{u}^{i}-\textbf{\emph{c}}_{v}^{i}\|_2+\lambda_{d}^{i}\cdot{|d_{u}^{i}-d_{v}^{i}|}}{\sigma^{2}}), & \text{if $r_v^i\in\Omega_{u}^{i}$}\\
0, & \text{otherwise}
\end{cases}
\end{equation}
where $\textbf{\emph{c}}_{u}^{i}$ and $d_{u}^{i}$ denote the mean L*a*b* color and depth value of superpixel $r^i_u$, $\Omega_{u}^{i}$ represents the neighbor set of superpixel $r^i_u$, $\|\cdot\|_2$ is the 2-norm of vector, $\lambda_{d}^{i}$ is the depth confidence measure, and $\sigma^{2}$ is a constant control parameter.\par
In our method, the seed superpixels are selected based on RGBD saliency value produced by depth propagation. Then, the top $\kappa$ superpixels with higher saliency values are considered as the saliency seeds, and the bottom $\kappa$ superpixels with lower saliency values are treated as the background seeds. In our experiments, $\kappa$ is set to 10. The initialized propagation score of the superpixel is defined as follow.
\begin{equation}
S_{0}^{i}(r_{n}^{i})=
\begin{cases}
1, & \text{if $r_{n}^{i}\in \Psi_F$}\\
0, & \text{if $r_{n}^{i}\in \Psi_B$}\\
S_{dp}^i(r_{n}^{i}), & \text{otherwise}
\end{cases}
\end{equation}
where $\Psi_F$ represents the saliency seed set, and $\Psi_B$ denotes the background seed set.\par
Using the labeled seeds, the saliency is propagated on the graph, and the score with saliency propagation is achieved by:
\begin{equation}
S_{sp}^{i}(r^i_m)=\sum_{n=1}^{N_i} w^i_{mn} \cdot S^{i}_0(r^i_n)
\end{equation}
where $w^i_{mn}$ is the element of the affinity matrix.\par

\subsection{Deletion Scheme}
The addition scheme is used to improve and optimize the saliency map from the perspective of intra-image. On the other hand, the inter-image information plays an important role in co-saliency detection. Therefore, a deletion scheme is designed to capture the corresponding relationship among multiple images, which aims to suppress the common and non-common backgrounds, and enhance the common salient regions from the perspective of multiple images. In our deletion scheme, a superpixel-level similarity measurement is constructed to represent the similarity relationship between two superpixels. Then, a common probability function using the similarity measurement is used to calculate the likelihood of each superpixel belonging to the common regions.\par
\subsubsection{\textbf{Multiple cues based similarity measurement}}
In deletion scheme, the color cue, depth cue, and saliency cue are combined into a measurement to evaluate the similarity between two superpixels.\par
\emph{RGB similarity.} The color histogram and texture histogram \cite{R47,R48} are used to represent the RGB feature on the superpixel level, which are denoted as $HC_m^i$ and $HT_m^i$, respectively. Then, the Chi-square measure is employed to compute the feature difference. Thus, the RGB similarity is defined as:
\begin{equation}
S_{c}(r^i_m,r^j_n)=1-\frac{1}{2}[\chi^2(HC_m^i, HC_n^j)+\chi^2(HT_m^i, HT_n^j)]
\end{equation}
where $r^i_m$ and $r^j_n$ are the superpixels in image $I^i$ and $I^j$, respectively, and $\chi^2(\cdot)$ denotes the Chi-square distance function.\par

\emph{Depth similarity.} Two depth consistency measurements, namely depth value consistency and depth contrast consistency, are composed of the final depth similarity measurement, which is defined as:
\begin{equation}
S_{d}(r^i_m,r^j_n)=\exp(-\frac{W_d(r^i_m,r^j_n)+W_c(r^i_m,r^j_n)}{\sigma^2})
\end{equation}
where $W_d(r^i_m,r^j_n)$ is the depth value consistency measurement to evaluate the inter image depth consistency, due to the fact that the common regions should appear similar depth values.
\begin{equation}
W_d(r^i_m,r^j_n)=|d^i_m-d^j_n|
\end{equation}
$W_c(r^i_m,r^j_n)$ describe the depth contrast consistency, because the common regions should represent more similar characteristic in depth contrast measurement.
\begin{equation}
W_c(r^i_m,r^j_n)=|D_c(r^i_m)-D_c(r^j_n)|
\end{equation}
with
\begin{equation}
D_c(r^i_m)=\sum_{k\neq m}|d^i_m-d^i_k|\exp(-\frac{\|\emph{\textbf{p}}_m^i-\emph{\textbf{p}}_k^i\|_2}{\sigma^2})
\end{equation}
where $D_c(r^i_m)$ denotes the depth contrast of superpixel $r^i_m$, $\emph{\textbf{p}}_m^i$ denotes the position of superpixel $r^i_m$, and $\sigma^2$ is a constant.\par

\emph{Saliency similarity. }Inspired by the prior that the common regions should appear more similar in single saliency map compared to other regions, the output saliency map from the addition scheme is used to define the saliency similarity measurement in our work:
\begin{equation}
S_s(r^i_m,r^j_n)=\exp(-|S_{sp}^i(r^i_m)-S_{sp}^j(r^j_n)|)
\end{equation}
where $S_{sp}^i(r^i_m)$ is the saliency score of superpixel $r^i_m$ based on Eq. (8).\par

\emph{Combination similarity.} Based on these cues, the combination similarity measurement is defined as the average of the three similarity measurements.
\begin{equation}
S_M(r^i_m,r^j_n)=\frac{S_c(r^i_m,r^j_n)+S_d(r^i_m,r^j_n)+S_s(r^i_m,r^j_n)}{3}
\end{equation}
where $S_c(r^i_m,r^j_n)$, $S_d(r^i_m,r^j_n)$, and $S_s(r^i_m,r^j_n)$ are the normalized RGB, depth, and saliency similarities between superpixel $r^i_m$ and $r^j_n$, respectively. A larger $S_M(r^i_m,r^j_n)$ value corresponds to greater similarity between two superpixels.\par

\subsubsection{\textbf{Common probability}}
For co-saliency detection, it is necessary to discriminate whether the selected salient objects are common or not. Thus, how to determine the common objects is a key point for co-saliency detection. In general, the common object is defined as the object with repeated occurrence in multiple images. Based on this definition, the common probability function is used to evaluate the likelihood that a superpixel belongs to the common regions, and it is defined as the sum of maximum matching probability among different images. For each superpixel $r^i_m$, only the most matching superpixel $r^j_k$ in image $I^j$ is selected for calculation, which is denoted as:
\begin{equation}
r^j_k=\arg \max_{n\in[1,N_j]} S_M(r^i_m,r^j_n)
\end{equation}
where $r^j_k$ is the most matching/similar superpixel in image $I^j$ for superpixel $r^i_m$ based on the maximum combination similarity score, and $N_j$ represents the number of superpixles in image $I^j$.\par
Then, these selected superpixels from different images are used to calculate the common probability:
\begin{equation}
P_c^i(r^i_m)=\frac{1}{N-1}\sum_{j=1,j\neq i}^{N} S_M(r^i_m,r^j_k)
\end{equation}
where $r^j_k$ is the most matching superpixel in image $I^j$ for superpixel $r_m^i$, and $N$ denotes the number of images in an image group. Finally, the updated co-saliency map of deletion scheme is denoted as:
\begin{equation}
S_{del}^i(r^i_m)=S_{sp}^i(r^i_m)\cdot P_c^i(r^i_m)
\end{equation}
where $S^i_{sp}(r^i_m)$ is the saliency score of superpixel $r_m^i$ produced by the addition scheme. Fig. \ref{fig3}(d) shows the co-saliency map after addition and deletion schemes. Compared with the initialized saliency map shown in Fig. \ref{fig3}(c), the co-salient object appears to be more consistency and the backgrounds are effectively suppressed.\par

\subsection{Iteration Scheme}
\begin{figure}[!t]
\centering
\includegraphics[width=1\linewidth]{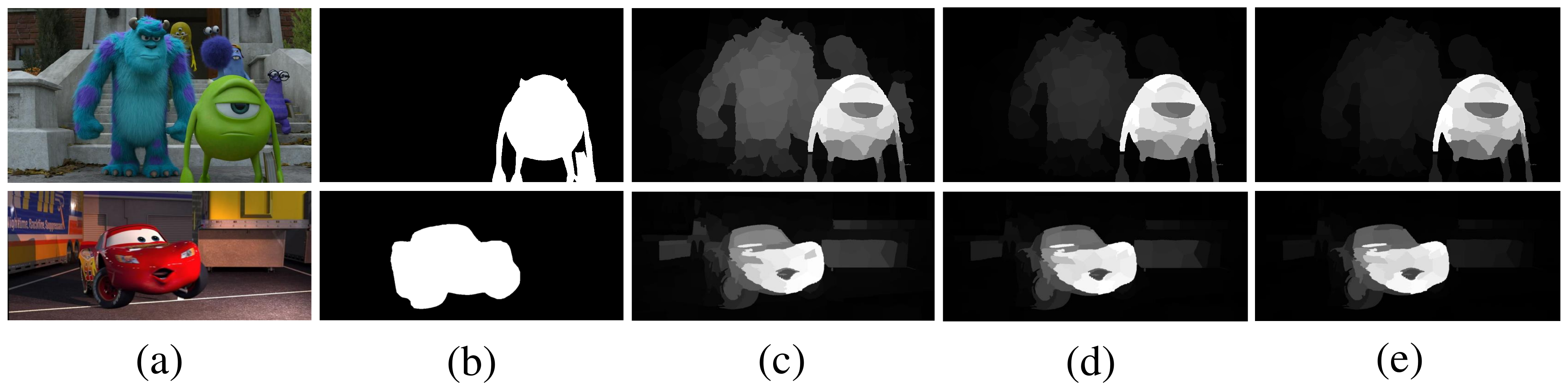}
\caption{Some examples of the iteration scheme. (a) RGB image. (b) Ground truth. (c) The initial saliency map through the addition and deletion scheme. (d) The saliency map after the first iteration. (e) The final saliency result. }
\label{fig5}
\end{figure}
In order to obtain more superior co-saliency map, an iterative scheme is designed in our framework, as shown in Fig. \ref{fig2}. The iterative scheme works as a refinement model to combine the addition and deletion steps and refine the co-saliency map in loop. In the iteration scheme, a heuristic termination strategy is set by checking the maximum iteration number $I_{max}$ and the difference between two iterations. Specifically, the second termination condition is introduced to check whether the saliency result becomes stable or not, which is formulated as the average difference between two iteration results.
\begin{equation}
D_t^i=(\frac{1}{\Pi}\sum |S_{del}^i(t)-S_{del}^i(t-1)|)\leq \zeta
\end{equation}
where $S_{del}^i(t)$ is the co-saliency map produced after the $t^{th}$ iteration optimization, $\Pi$ represents the number of pixels in the co-saliency map, and $\zeta$ is a given threshold to determine whether the iteration should be terminated or not, which is set to 0.1 in all experiments. Until $D_t^i\leq \zeta$, the iteration will be terminated and output the final co-saliency map, otherwise, the iteration will continue. Some visual examples of the iteration scheme are shown in Fig. \ref{fig5}. The third column shows the original co-saliency result, and the first iteration and the final co-saliency maps are shown in the last two columns of Fig. \ref{fig5}. From the figure, we can see that the initial co-saliency map is improved obviously with the iteration processing. For example, the cartoon with blue hair (named Sulley) is suppressed effectively since it is not a common object in the image group. Similarly, the background regions around the red car are also suppressed through the iteration scheme. The overall framework of the proposed method is summarized in Algorithm 1.

\begin{algorithm}[htb]
  \caption{The Overall Framework.}
  \begin{algorithmic}[1]
    \Require
      The RGB images and depth maps in an image group.
    \Ensure
      The co-saliency map for each image.
    \For {each image in the group}
         \State Obtain the initialized saliency map using Eq. (1);
         \Repeat
            \State Conduct the addition scheme using Eqs. (2-8);
            \State Conduct the deletion scheme using Eqs. (9-18);
         \Until {$D_t^i\leq \zeta$ or $t\geq I_{max}$}
    \EndFor
  \end{algorithmic}
\end{algorithm}

\section{Experiments}
In this section, the proposed RGBD co-saliency framework is evaluated on two RGBD co-saliency datasets. The qualitative and quantitative comparison with other state-of-the-art methods are presented. In addition, the analysis and discussion are conducted, which include the analysis of each module in the framework and the discussion of one-for-one option co-saliency framework.\par
\begin{figure*}[!t]
\centering
\includegraphics[width=1\linewidth]{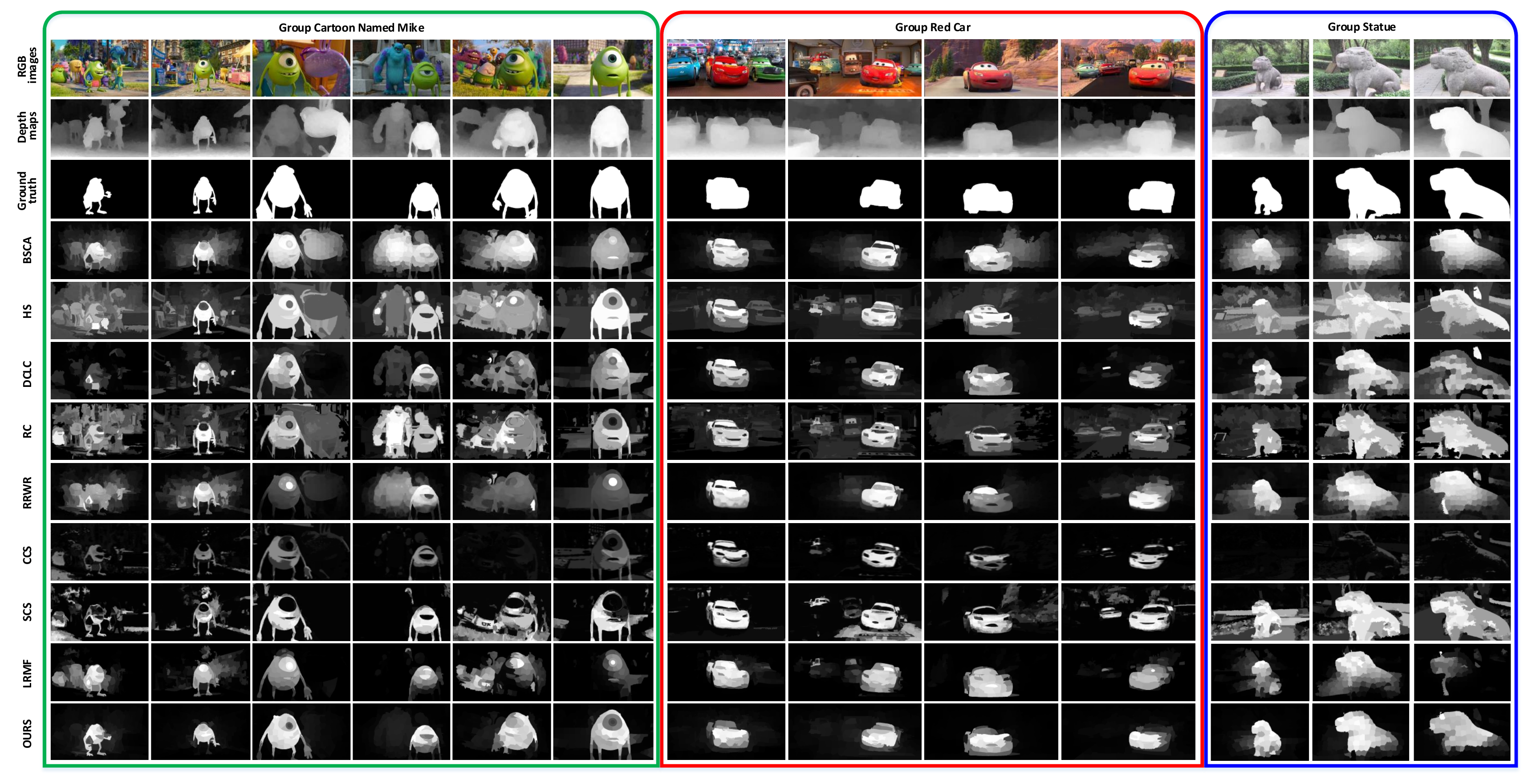}
\caption{Visual comparison of different saliency and co-saliency detection methods on RGBD Cosal150 dataset. }
\label{fig6}
\end{figure*}

\begin{figure*}[!t]
\centering
\includegraphics[width=1\linewidth]{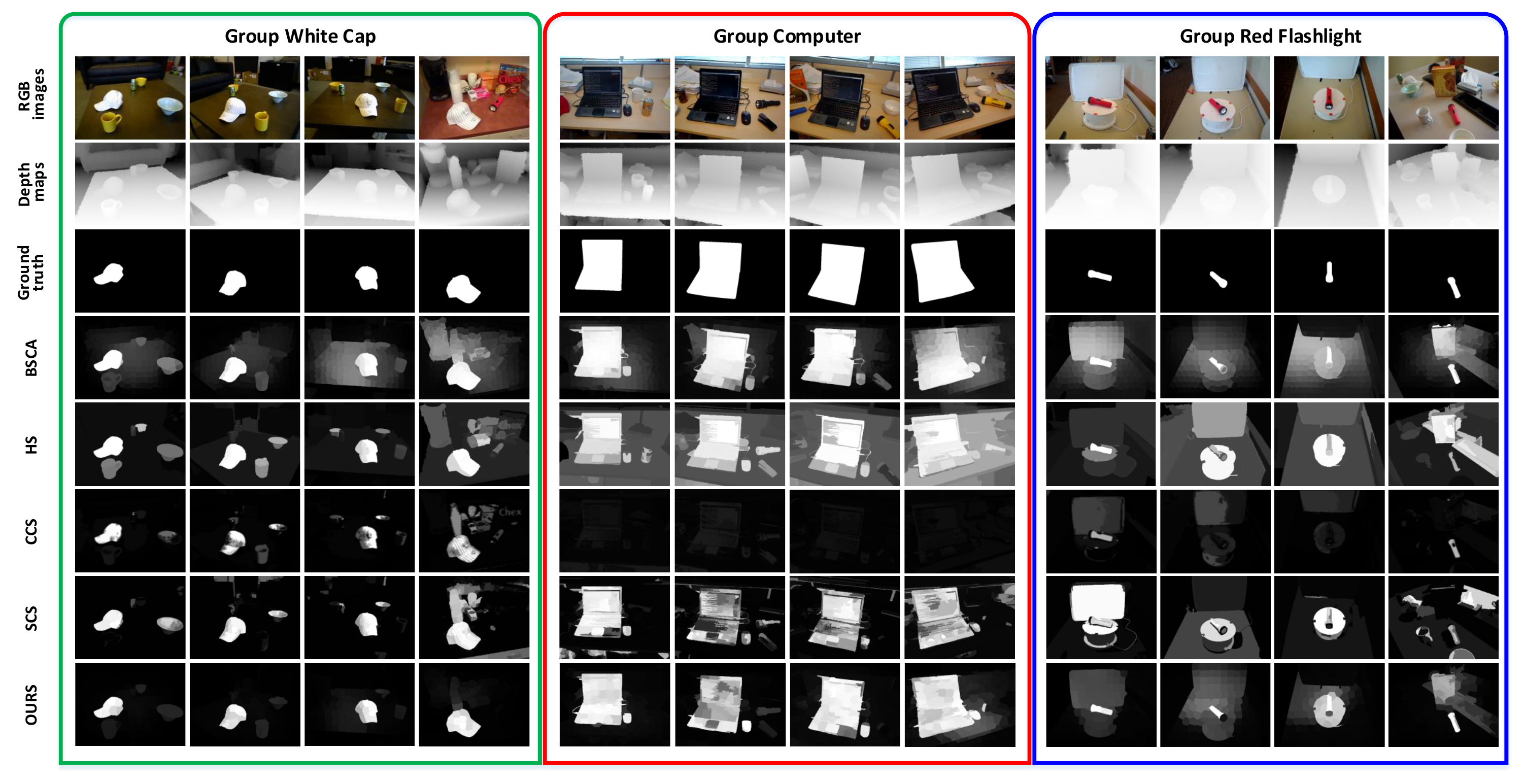}
\caption{Visual comparison of different saliency and co-saliency detection methods on RGBD Coseg183 dataset. }
\label{fig7}
\end{figure*}

\begin{figure*}[!t]
\centering
\includegraphics[width=1\linewidth]{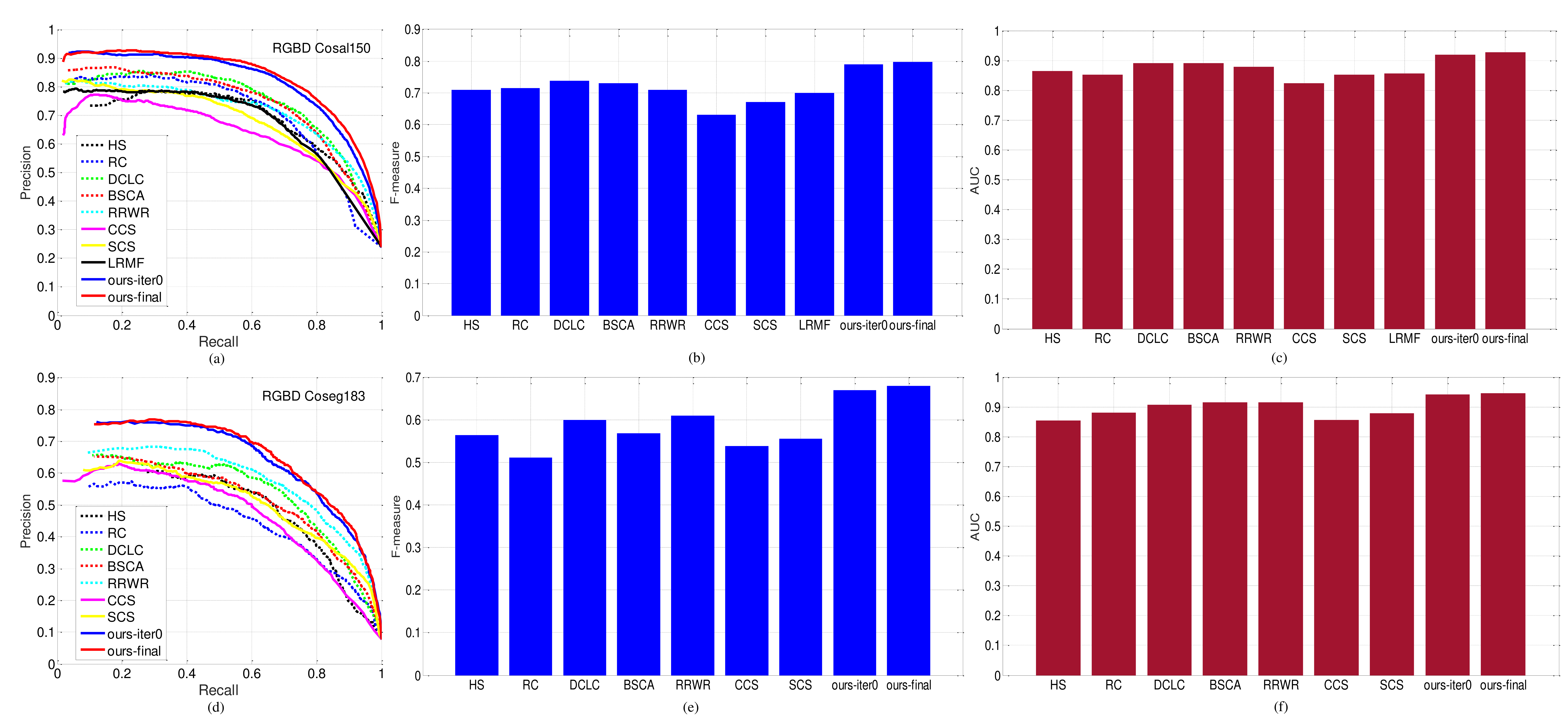}
\caption{Quantitative comparisons between the proposed method and the state-of-the-art methods on two datasets. Notice that ``ours-iter0'' means the co-saliency without iteration scheme, and ``ours-final'' denotes the co-saliency result with iteration scheme. (a)-(c) PR curves, F-measure and AUC scores on RGBD Cosal150 dataset. (d)-(f) PR curves, F-measure and AUC scores on RGBD Coseg183 dataset. }
\label{fig8}
\end{figure*}
\subsection{Experimental Settings}
The proposed co-saliency framework is evaluated on two RGBD benchmarks: the RGBD Coseg183 dataset\footnote{ \url{http://hzfu.github.io/proj_rgbdseg.html} } \cite{R23} and the RGBD Cosal150 dataset\footnote{  \url{https://rmcong.github.io/proj_RGBD_cosal.html}}. The RGBD Coseg183 dataset contains 183 images with pixel-level ground-truth that are distributed in 16 indoor scenes. For more comprehensive comparison and analysis, we collect 21 image sets containing totally 150 images from RGBD NJU-1985 dataset \cite{R36} for RGBD co-saliency detection, which is called RGBD Cosal150 dataset. Pixel-level ground-truth for each image is manually labeled in the dataset.\par
Three quantitative criteria are adopted to evaluate the co-saliency map, which include the Precision-Recall (PR) curve, F-measure, and AUC score. The precision and recall score are computed by thresholding the saliency map into a binary map, and comparing the binary map against the ground truth. The PR curve demonstrates the relationship between precision and recall of saliency map at different thresholds. F-measure \cite{R49} is defined as the weighted mean of precision $P$ and recall $R$, which is denoted as:\par
\begin{equation}
F_{\beta}=\frac{(1+\beta^{2})P\times R}{\beta^{2}\times P+ R}
\end{equation}
where $\beta^{2}$ is set to 0.3 that emphasizes the precision more than recall. In addition, AUC evaluates the object detection performance, which is computed as the area under the standard ROC curve. In the proposed method, the number of superpixels for each image is set to 200, the maximum iteration number is set to 5 for balancing the computational complexity and performance, and the method is implemented in MATLAB 2014a on a Quad Core 3.5GHz workstation with 16GB RAM. The proposed method costs average 42.67 seconds to process one image. The project is available on our website\footnote{  \url{https://rmcong.github.io/proj_RGBD_cosal_tcyb.html} }.\par

\subsection{Comparison with State-of-the-art Methods}

In this section, we compare the proposed method with 8 state-of-the-art methods, which include RC \cite{R8}, DCLC \cite{R9}, RRWR \cite{R10}, HS \cite{R11}, BSCA \cite{R12}, CCS \cite{R18}, SCS \cite{R30}, and LRMF \cite{R31}. The first five single image saliency methods are regarded as the input of the proposed framework, and the last three methods are the state-of-the-art co-saliency methods.\par
\begin{figure*}[!t]
\centering
\includegraphics[width=1\linewidth]{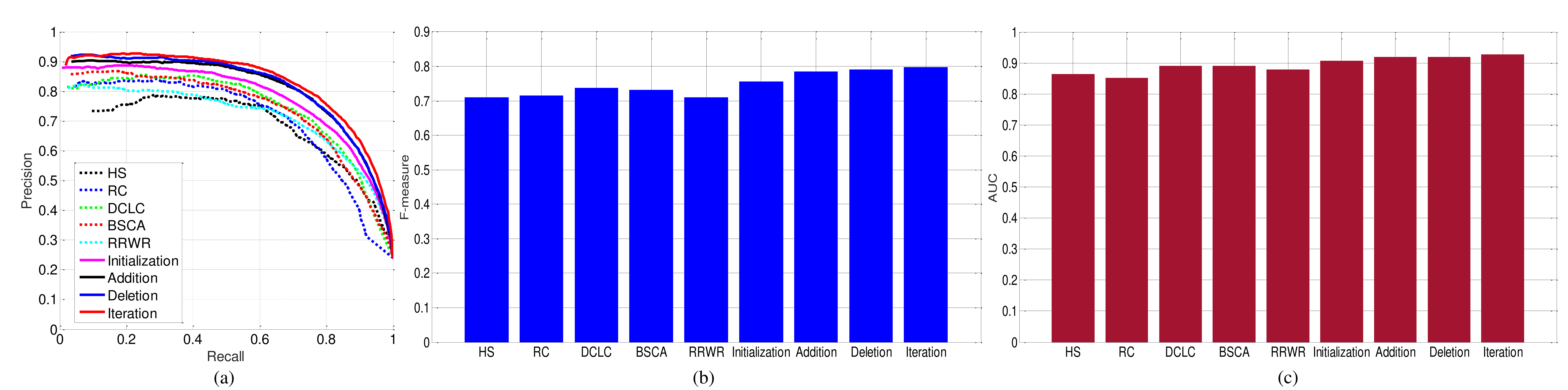}
\caption{Quantitative comparisons of the each part of the proposed framework on RGBD Cosal150 dataset. (a) PR curves. (b) F-measure. (c) AUC scores. }
\label{fig9}
\end{figure*}
For subjective evaluation, the visual examples on two datasets are shown in Figs. \ref{fig6} and \ref{fig7}, which consist of three image groups on RGBD Cosal150 dataset, i.e., the group of cartoon named Mike, red car, and statue, as well as three groups on RGBD Coseg183 dataset, i.e., the group of white cap, computer, and red flashlight. From Fig. \ref{fig6}, we can see that the single image saliency methods (e.g., RC, HS, and RRWR) fail to discover the co-salient objects effectively and accurately. Taking the group Mike as an example, the common salient object is the green cartoon with big eye. However, many non-common objects, such as the cartoon with blue hair and the purple snake, are detected as the salient objects in the single saliency models. In addition, some background regions are not effectively suppressed in groups red car and statue, such as the trees and non-salient cars. In a word, the single saliency detection methods fail to detect the common salient objects in co-saliency scenarios. Therefore, it is essential that a co-saliency framework should be designed to convert the single saliency map into co-saliency result. The co-saliency map produced by our framework is shown in the last row of Fig. \ref{fig6}. Compared with the single saliency maps, the common salient regions are highlighted more consistent and accurate, and the backgrounds are suppressed effectively. To further evaluate the proposed method, three state-of-the-art co-saliency methods are introduced for comparison. From the figures, it indicates that the proposed approach can effectively highlight the common salient regions from the image group, and robustly suppress the background regions even when the salient regions exhibit large variations in shape and direction or the background is very complex and interferential. In contrast, the RGBD Coseg183 dataset is more difficult and challenging for co-saliency detection, and some visual examples are shown in Fig. \ref{fig7}. The proposed method achieves better performance compared with the other saliency and co-saliency detection methods. For example, the non-common objects, e.g. the white bowl and yellow cup, are effectively suppressed in the white cap group compared to other methods. Moreover, in computer group, the consistency and homogeneity of the salient object is improved obviously compared with others. In red flashlight group, the red flashlight using our method is highlighted more effective than others. However, some backgrounds are still retained in the final result due to the small size and complex scene.\par
The quantitative comparison results including the PR curves, F-measure, and AUC scores are reported in Fig. \ref{fig8}. As can be seen, on the RGBD Cosal150 dataset, the proposed method achieves the highest precisions of the whole PR curves, the largest F-measure and AUC score compared with other methods. The same conclusion can be drawn from the results on RGBD Coseg183 dataset. From the PR curves on both two datasets, it can be seen that the final co-saliency result (the red line) reaches the highest level in all curves, and the performance of co-saliency framework is obviously superior to the five original single image saliency models. It also demonstrates that the proposed co-saliency framework achieves the goal of converting the single saliency results into co-saliency scenarios. The F-measure and AUC scores also support the conclusion. In the proposed RGBD co-saliency detection framework, we aim to design a many-for-one structure, i.e. multiple single saliency maps input and one co-saliency map output, to synthesize the superiority of different single saliency maps. In order to prove the effectiveness and versatility of the proposed algorithm, another one-for-one option is also implemented and evaluated, and the relevant results will be discussed in Section IV-E.\par
\subsection{Module Analysis}
\begin{figure}[!t]
\centering
\includegraphics[width=1\linewidth]{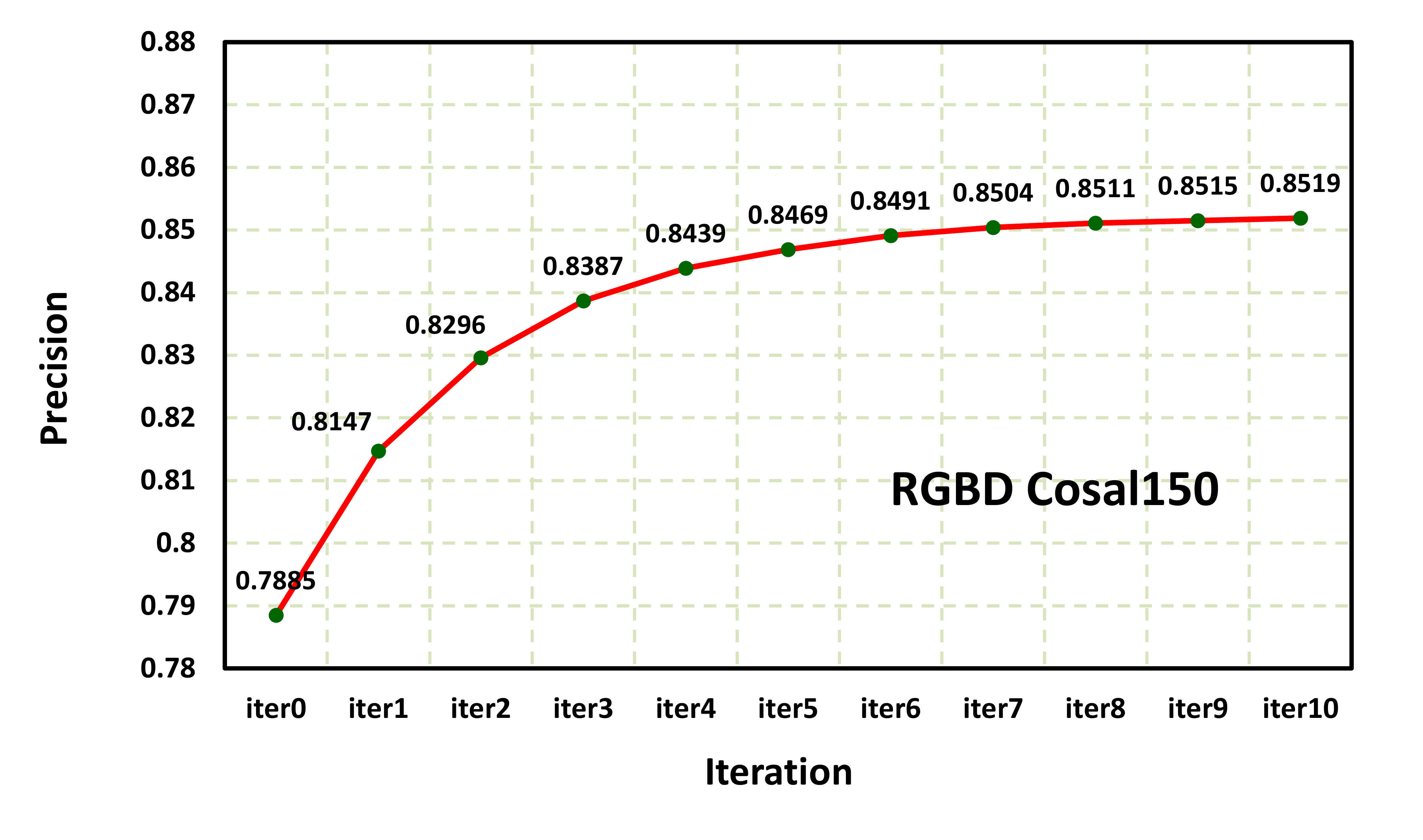}
\caption{The average precision of each iteration on RGBD Cosal150 dataset. }
\label{fig10}
\end{figure}

In this section, we comprehensively evaluate each module of the proposed framework including the initialization, addition scheme, deletion scheme, and iteration scheme. The quantitative comparisons on RGBD Cosal150 dataset are shown in Fig. \ref{fig9}, and the evaluation result of the iteration scheme is represented in Fig. \ref{fig10}. In the initialization process, the original five saliency maps are integrated to produce a baseline for co-saliency detection, and its PR curve is marked as carmine in Fig. \ref{fig9}(a). Compared with the PR curves and F-measure of the original single saliency results, it indicates that the initialization result achieves better performance and produces a preferable baseline for later co-saliency detection. In addition scheme, depth shape prior is proposed to introduce the depth information into the framework, and label propagation is used to further optimize the saliency result. As shown in Fig. \ref{fig9}, the saliency map through the addition scheme (the black line in PR curves) is improved significantly compared to the initialization result (the carmine line), and the F-measure and AUC score also achieve higher scores. Then, the deletion scheme is conducted to introduce the inter-image corresponding information into the framework and produce the initial co-saliency map. All the quantitative measurements in Fig. \ref{fig9} show that the initial co-saliency result without iteration scheme obtains the best performance compared to other modules.\par

\begin{figure*}[htbp]
\centering
\includegraphics[width=0.95\linewidth]{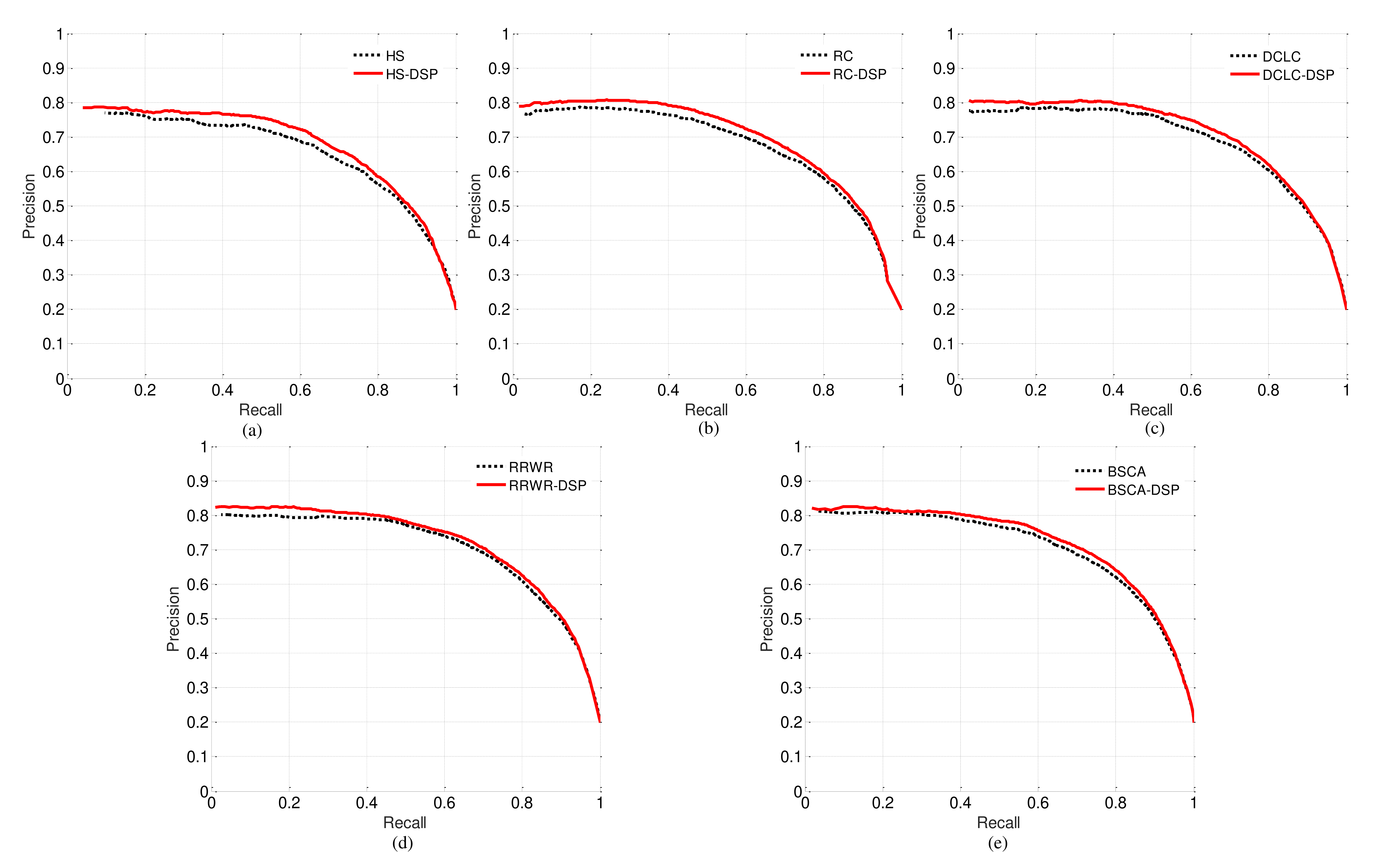}
\caption{Quantitative evaluation of the depth shape prior on RGBD400 dataset. The black line in each PR curve denotes the original RGB saliency result, and the red line represents the saliency result with depth shape prior. The PR curves for (a) HS, (b) RC, (c) DCLC, (d) RRWR, and (e) BSCA methods. }
\label{fig11}
\end{figure*}

In addition, an iteration scheme is designed to further update the co-saliency map and achieve more consistent result. The PR curves in Fig. \ref{fig9}(a) demonstrate that the performance is obviously improved using the iteration scheme, in which the blue line denotes the initial co-saliency result and the red line represents the final co-saliency result with iteration scheme. With the conduct of the iteration scheme, the performance of co-saliency detection is continually optimized according to the F-measure and AUC score. In other words, all of these measurements illustrate the effectiveness of the iterative mechanism proposed in our framework. In order to verify the rationality of the iteration termination condition, an experiment of ten iterations without termination conditions are conducted on RGBD Cosal150 dataset, and the detailed quantitative comparison results are shown in Fig. \ref{fig10}. From the average precision curve, we can see that, with the iterative progress, the performance of the algorithm tends to be stable gradually. In general, the termination conditions will not be satisfied after the first iteration, and its improved level is most noticeable. Moreover, most of the images will satisfy the termination condition after 3-4 iterations, that is, the co-saliency map no longer appears obvious changes. Thus, it also demonstrates that the maximum iteration number of 5 is reasonable in the experiments.\par

\subsection{Evaluation of the Depth Shape Prior}
\begin{table}[!t]
\renewcommand\arraystretch{1.8}
\centering
\caption{The F-measure of the depth shape prior on the RGBD400 Dataset}
\begin{tabular}{c|c|c|c|c|c}
\toprule[1.2pt]
 & HS & RC & DCLC & RRWR & BSCA \\[0.5ex]
\hline
Without DSP & $0.6661$ & $0.6732$ & $0.6914$ & $0.7040$ & $0.7022$ \\
\hline
With DSP & $0.6904$ & $0.6914$ & $0.7094$ & $0.7132$ & $0.7136$ \\
\hline
Percentage Gain & $3.65\%$ & $2.70\%$ & $2.60\%$ & $1.31\%$ & $1.62\%$ \\
\bottomrule[1.2pt]
\end{tabular}
\label{tab1}
\end{table}

\begin{table}[!t]
\renewcommand\arraystretch{1.8}
\centering
\caption{The F-measure of the depth shape prior on the RGBD Cosal150 Dataset}
\begin{tabular}{c|c|c|c|c|c}
\toprule[1.2pt]
 & HS & RC & DCLC & RRWR & BSCA \\[0.5ex]
\hline
Without DSP & $0.7101$ & $0.7163$ & $0.7385$ & $0.7106$ & $0.7318$ \\
\hline
With DSP & $0.7294$ & $0.7457$ & $0.7642$ & $0.7294$ & $0.7502$ \\
\hline
Percentage Gain & $2.72\%$ & $4.10\%$ & $3.48\%$ & $2.65\%$ & $2.51\%$ \\
\bottomrule[1.2pt]
\end{tabular}
\label{tab2}
\end{table}

\begin{figure*}[!t]
\centering
\includegraphics[width=0.95\linewidth]{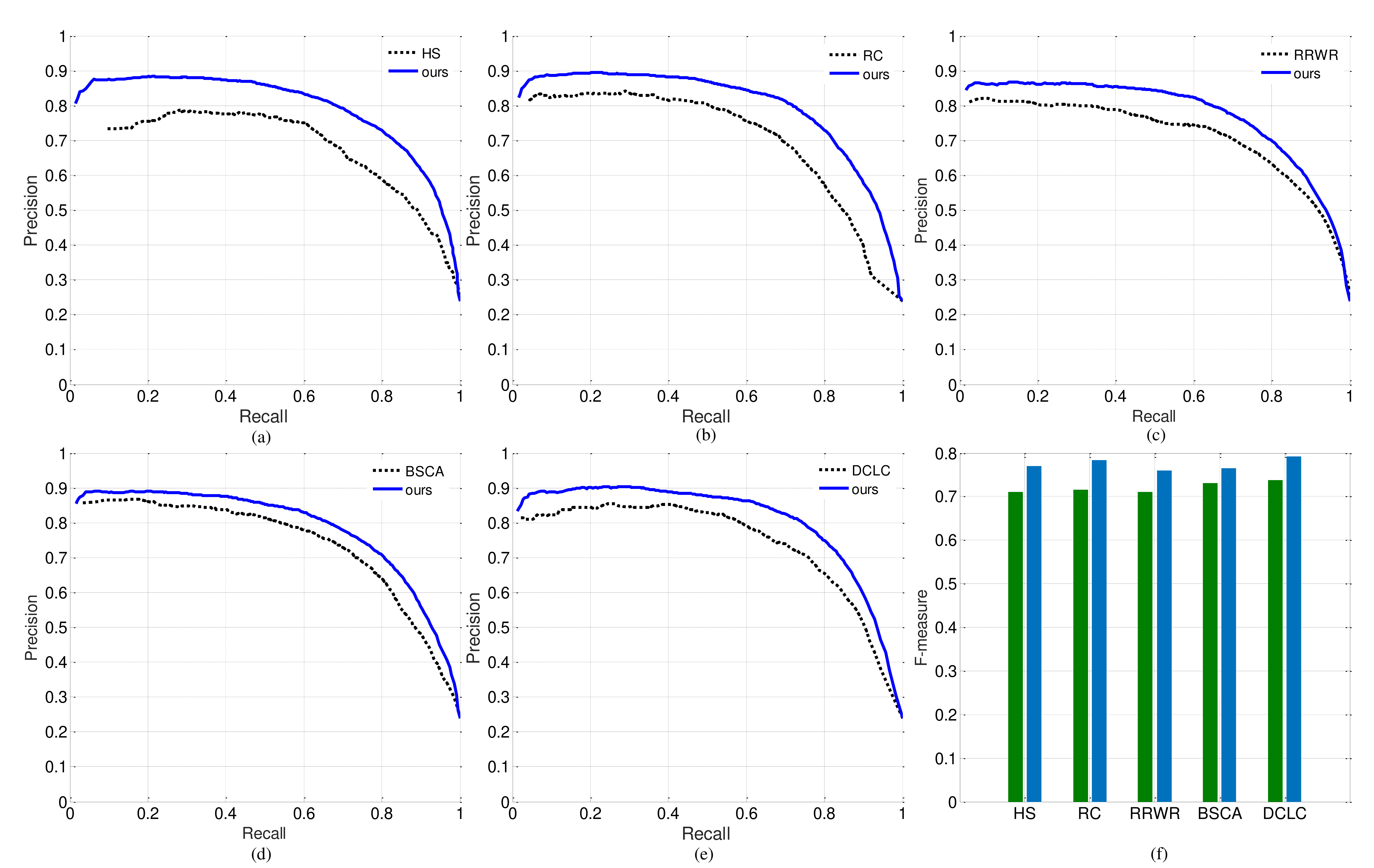}
\caption{Quantitative evaluation of one-for-one option for our framework on RGBD Cosal150 dataset. The black line in each PR curve denotes the original RGB saliency result, and the blue line represents the final co-saliency result using the proposed framework. The PR curves with different input saliency maps, i.e., (a) HS, (b) RC, (c) RRWR, (d) BSCA, and (e) DCLC. (f) F-measure of the one-for-one framework. }
\label{fig12}
\end{figure*}

In the framework, a novel depth descriptor, namely depth shape prior (DSP), is proposed to introduce the depth information to assist the identification of the co-salient objects. Introducing the depth shape prior into RGB saliency model, the 2D saliency model will turn into a RGBD saliency model and achieve a better performance. In this section, we evaluate the performance of depth shape prior on RGBD400 saliency dataset \cite{R36}, and the relevant results are shown in Fig. \ref{fig11} and TABLE \ref{tab1}. Five different 2D saliency maps produced by BSCA, RC, HS, RRWR, and DCLC methods are used as the original saliency maps. In PR curves, the black line denotes the original RGB saliency result, and the red line represents the saliency result with depth shape prior. From the PR curves shown in Fig. \ref{fig11}, it can be seen that the saliency result with depth shape prior achieves the higher precisions of the whole PR curves compared to the 2D saliency results, and the F-measure also arrives at the consistent conclusion from TABLE \ref{tab1}. For the F-measure, the maximum percentage gain achieves 3.65\% for HS method, and the average percentage gain achieves 2.38\%. In order to further illustrate the effectiveness of DSP descriptor in our model, we have conducted an experiment on the RGBD Cosal150 dataset, and the results are shown in TABLE \ref{tab2}. From the table, it can be seen that the F-measure achieves the maximum percentage gain of 4.10\% for RC method, and the average percentage gain achieves 3.09\%. These experiments demonstrated that the depth information could improve the performances of the co-saliency. In other words, the depth shape prior can be used as an independent descriptor that converts the 2D saliency map into RGBD saliency map.\par

\subsection{Discussion}
The proposed RGBD co-saliency detection framework is designed as a many-for-one model, that is, multiple single 2D saliency maps input and one RGBD co-saliency map output. In fact, our framework can also achieve one-for-one model. In other words, if there is only one saliency map is embedded into the framework, it can also output one RGBD co-saliency map. The experimental comparison is reported in Fig. \ref{fig12}. The PR curves and F-measure demonstrate that the one-for-one option also achieves the transformation from single image saliency map to RGBD co-saliency map, and obtains better performance of co-saliency detection. In general, the better the saliency map is, the better the co-saliency map achieves. This is, of course, the reason why the multiple saliency maps are fused at first in the proposed framework. It can provide a better baseline for later detection in order to achieve more accurate and stable co-saliency result. However, as the results shown in Fig. \ref{fig12}, the proposed framework can also acquire satisfying result when only one saliency map is embedded.\par
\begin{figure}[!t]
\centering
\includegraphics[width=1\linewidth]{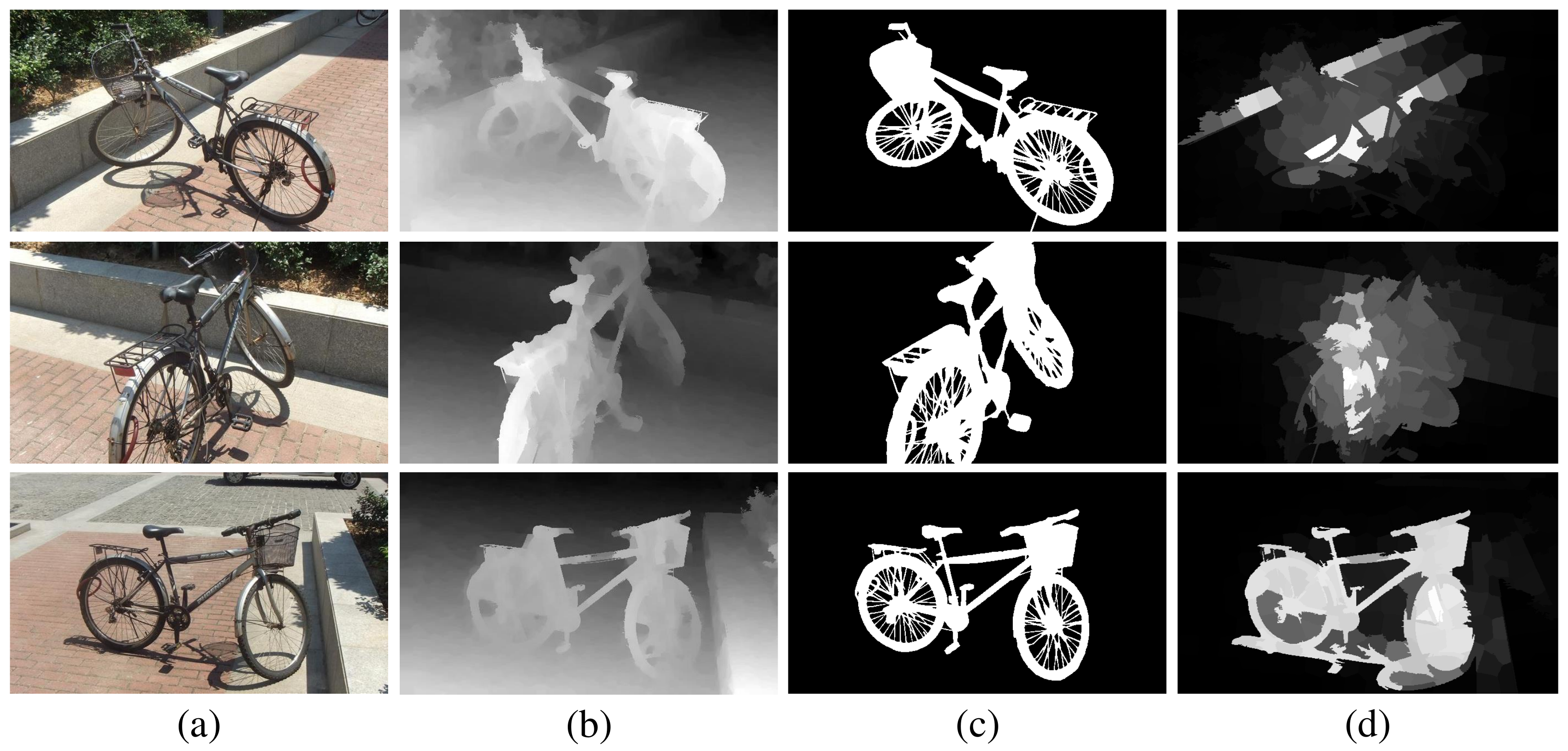}
\caption{Some challenging examples for our RGBD co-saliency detection model. (a) RGB image. (b) Depth map. (c) Ground truth. (d) The co-saliency result produced by our framework. }
\label{fig13}
\end{figure}
In addition, we discuss some degenerated cases of the proposed RGBD co-saliency framework in this section. The details are shown in Fig. \ref{fig13}. In this group, there are lots of textures and detail regions, such as the spokes and back seat of the bike in each image. These regions are difficult to detect completely and accurately for our framework. The main reason is that our co-saliency model focuses on low-level features extraction without learning, which could not capture the details very well. In the future, the low-level traditional features and the high-level learning features can be integrated to further describe the object characteristics of the image.\par

\section{Conclusion}
In this paper, an iterative RGBD co-saliency framework is proposed to convert the existing 2D saliency model into RGBD co-saliency scenario. Three schemes are integrated into the framework, which include addition scheme, deletion scheme and iteration scheme. The addition scheme is used to optimize the single saliency map and introduce the depth information into the framework using a novel descriptor named depth shape prior. The deletion scheme aims at capturing the inter-image constraints and suppressing the non-common regions using a common probability function, which is formulated as the likelihood of each superpixel belonging to the common regions. Finally, an iterative scheme is designed to obtain more homogeneous and consistent co-saliency map. The comprehensive comparison and discussion on two RGBD co-saliency datasets have demonstrated that the proposed method outperforms other state-of-the-art saliency and co-saliency models.\par

\ifCLASSOPTIONcaptionsoff
  \newpage
\fi

\end{document}